\documentclass[letterpaper, 10 pt, conference]{ieeeconf}  % Comment this line out if you need a4paper

\IEEEoverridecommandlockouts                              % This command is only needed if 
\usepackage{graphicx} % for pdf, bitmapped graphics files
\graphicspath{ {./Figures/} }
\usepackage{amsmath} % assumes amsmath package installed
\usepackage{multicol}
\usepackage{cite}
\usepackage{multirow}
\usepackage{booktabs}

\title{\LARGE \bf
Exploring Stiffness Gradient Effects in Magnetically Induced Metamorphic Materials via Continuum Simulation and Validation
}

\author{Wentao Shi$^{\dag}$, Yang Yang$^{\dag}$, Yiming Huang and Hongliang Ren*% <-this % stops a space
\thanks{*Corresponding author: hren@cuhk.edu.hk}
\thanks{$^{\dag}$These authors contributed equally to this work.}% <-this % stops a space
\thanks{This work was supported by Hong Kong Research Grants Council Collaborative Research Fund (CRF C4026-21G); General Research Fund (GRF) 14216022 and (GRF) 14204524; Research Grants Council (RGC)-NSFC/RGC Joint Research Scheme N\_CUHK420/22; and Shenzhen-Hong Kong-Macau Technology Research Programme (Type C) STIC Grant202108233000303.}
}

\begin{document}

\maketitle
\thispagestyle{empty}
\pagestyle{empty}

%%%%%%%%%%%%%%%%%%%%%%%%%%%%%%%%%%%%%%%%%%%%%%%%%%%%%%%%%%%%%%%%%%%%%%%%%%%%%%%%
\begin{abstract}

Magnetic soft continuum robots are capable of bending with remote control in confined space environments, and they have been applied in various bioengineering contexts. As one type of ferromagnetic soft continuums, the Magnetically Induced Metamorphic Materials (MIMMs)-based continuum (MC) exhibits similar bending behaviors. Based on the characteristics of its base material, MC is flexible in modifying unit stiffness and convenient in molding fabrication. However, recent studies on magnetic continuum robots have primarily focused on one or two design parameters, limiting the development of a comprehensive magnetic continuum bending model. In this work, we constructed graded-stiffness MCs (GMCs) and developed a numerical model for GMCs' bending performance, incorporating four key parameters that determine their performance. The simulated bending results were validated with real bending experiments in four different categories: varying magnetic field, cross-section, unit stiffness, and unit length. The graded-stiffness design strategy applied to GMCs prevents sharp bending at the fixed end and results in a more circular curvature. We also trained an expansion model for GMCs' bending performance that is highly efficient and accurate compared to the simulation process. An extensive library of bending prediction for GMCs was built using the trained model.

\end{abstract}

%%%%%%%%%%%%%%%%%%%%%%%%%%%%%%%%%%%%%%%%%%%%%%%%%%%%%%%%%%%%%%%%%%%%%%%%%%%%%%%%
\section{Introduction}

\begin{figure*}[h!]
    \centering
%       \framebox{\parbox{3in}{We suggest that you use a text box to insert a graphic (which is ideally a 300 dpi TIFF or EPS file, with all fonts embedded) because, in an document, this method is somewhat more stable than directly inserting a picture.
% }}
    \includegraphics[width=1.0\textwidth]{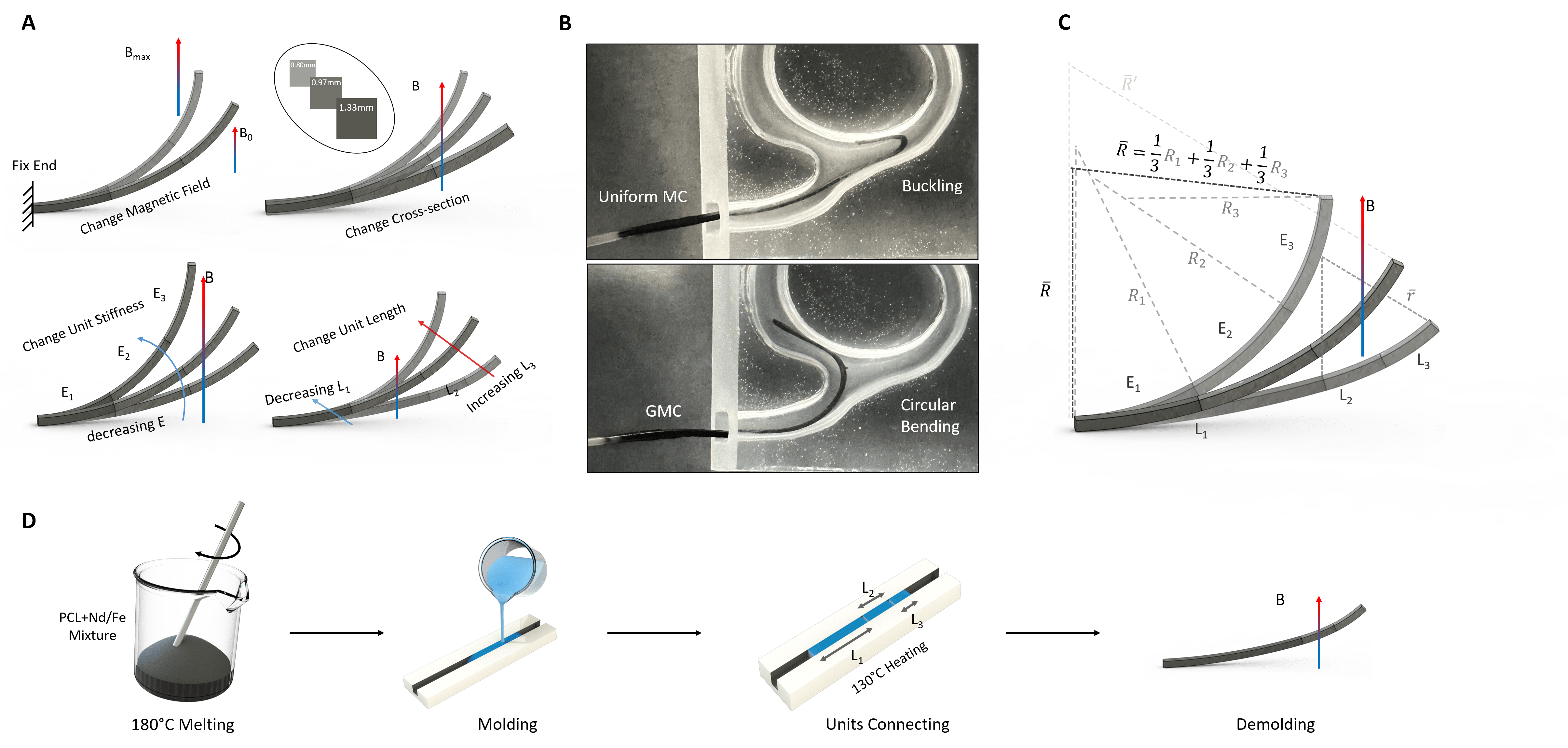}
    \caption{\textbf{GMC design strategy and fabrication.} (A) Demonstration of graded-stiffness MC (GMC)'s bending performances when varying external magnetic field, cross-section side length, unit stiffness,  and unit length. (B) Different bending behaviors of a uniform MC and a GMC for the same navigation task. (C) Modifying a specific design parameter will have a significant influence on the continuum's bending radius. (D) Molding fabrication of a GMC.}
    \label{figure1}
\end{figure*}

Magnetic soft robots have been widely studied in recent years and applied in various biomedical scenarios \cite{c1, c2}. Leveraging the characteristics of ferromagnetic materials, magnetic soft robots can respond to external magnetic fields, enabling diverse remote control strategies for navigation and actuation. Among the various shapes of magnetic soft robots, magnetic continuum robots are promising for bioengineering applications due to their small scale and simple structure. Researchers have demonstrated the potential of remotely controllable magnetic continuum robots in applications such as aneurysm treatments, epiretinal membrane surgeries, and vascular ablations \cite{c2, c3, c4,c5}.

As magnetic continuum robots benefit surgeons with remotely controllable minimally invasive operations, it is crucial to obtain an adequate and comprehensive understanding of how the parameters of these robots influence their performance. However, recent studies have typically focused on only one design parameter at a time. Specifically, for continuum robots with embedded permanent magnets, researchers primarily focused on the orientation of the embedded magnets \cite{c6, c7}. Additionally, the stiffness-changing characteristic was explored using a low-melting-point alloy as the body material for permanent magnet continuum robots \cite{c8}. For ferromagnetic continuum robots, greater attention has been paid to the continuum's magnetization profile, external magnetic field \cite{c9}, and the unit length \cite{c1, c10}. 
% These narrow perspectives hinder the comprehensive understanding of the movement of magnetic continuum robots.
While these perspectives provide unique insights into the movement, a comprehensive model is needed to present continuum performance from various aspects.

To address the gap, we considered four different parameters for this prediction and simulation problem: 1) magnetic field, 2) cross-section area, 3) unit stiffness, 4) unit length (Fig.~\ref{figure1}.A). To ensure that all parameters remain tunable and to facilitate a convenient fabrication process, we used magnetically induced metamorphic materials (MIMMs) \cite{c11} from our previous work to mold the ferromagnetic continuum. We conducted real-world bending tests of GMCs using a movable external permanent magnet (EPM) system. As the bending experiments continued, we observed that the GMC with a uniform structure tended to bend with a more pronounced curvature at its root (fixed end). Therefore, we introduced a graded-stiffness strategy for the GMC to achieve a more desired circular bending behavior. A GMC was divided into three sections with decreasing stiffness from the fixed end to the top. A numerical model was also constructed using COMSOL to predict the behaviors of GMCs under a magnetic field. 
% The comparison results between real experiments and COMSOL simulations over 20 cases indicate that we can predict the real bending behaviors of GMC from COMSOL simulations with accurate parameter inputs. 
The numerical results were further trained using a neural network to expand the prediction library, enabling accurate predictions of GMCs' bending movements without the need for time-consuming simulations. 
% As a supplement and expansion of current ferromagnetic continuum simulations, our work presents a substantial bending prediction library for GMC, enabling the selection of suitable GMC for various applications.

% [INTRODUCE THE IDEA OF GRADED STIFFNESS]
The core contributions of this work are listed as follows:
\begin{itemize}
    \item A new design strategy of the graded-stiffness structure is introduced for GMCs to achieve circular bending behavior.
    \item The numerical simulation of GMCs' bending performance is conducted. The simulation results are validated against real bending experiments.
    \item An accurate and efficient expansion model is trained to build a prediction library for GMCs' bending performance.
\end{itemize}
   
\section{Design Methodology and Fabrication}

\subsection{Design Methodology of GMC}

% Version with graded stiffness description
As one of the ferromagnetic soft continuum robots, the MC can perform magnetic field-induced bending due to the presence of neodymium magnetic nanoparticles. Numerous factors affect the bending performance of the ferromagnetic soft continuum. Based on related research, these factors include external magnetic fields, the magnetization profiles of the continuum, the stiffness of the continuum, and the unit length of each continuum section. However, the characteristics of the continuum's base material have previously hindered researchers from investigating all these parameters to present a comprehensive model. The physical and thermoelastic properties of Polycaprolactone (PCL, base material) enable us to mold the continuum easily with different cross-sections, unit stiffness, and unit lengths. In preliminary bending experiments, we noted that a normal structure MC bent sharply at its fixed end. The phenomenon is consistent with numerical results from a previous study \cite{c9}. This bending behavior led to undesired buckling in the navigation tests (Fig.~\ref{figure1}.B). To solve this problem, we proposed the graded-stiffness MC (GMC) with a more circular bending performance. Specifically, we mixed two PCL powders with different stiffness in various ratios for different parts: the rigid part for the bottom section, the transition part for the middle section, and the soft part for the top section, creating a graded-stiffness structure. The graded-stiffness structure ensures the transmission of forces and moments in the GMC and avoids the occurrence of buckling in navigation tasks. Furthermore, we mixed different stiffness base materials with various weight ratios of magnetic nanoparticles to create seven types of GMCs. With preset mixture ratios and silicone molding processes, different GMCs will have diverse cross-sections, stiffness, and lengths.

% HISTORY VERSION
% As one of the ferromagnetic soft continuum robots, GMC can perform magnetic field-induced bending due to the presence of neodymium magnetic nanoparticles. Numerous factors affect the bending performance of the ferromagnetic soft continuum. Based on related research, these factors include external magnetic fields, the magnetization profiles of the continuum, the stiffness of the continuum, and the unit length of each continuum section (Fig.~\ref{figure1}.C). However, the characteristics of the continuum's base material have previously hinder researchers from investigating all these parameters to present a comprehensive model. The physical and thermoelastic properties of PCL enable us to easily mold the continuum with different cross-sections, unit stiffness, and unit length. Specifically, we mixed two different stiffness PCL powders in various ratios for the fabrication of different parts: the harder part for the bottom section, the transition part for the middle section, and the softer part for the top section, creating a graded-stiffness structure. 
% After obtaining the graded-stiffness base materials, we mixed them with different weight ratios of magnetic nanoparticles to create seven types of GMC. With a preset mixture ratio and silicone mold, different GMC will have a unique cross-section, stiffness, and length.

In this work, we focused on four different parameters that have significant influences on the bending behaviors of GMCs, including the external magnetic field, GMC's cross-section side length, GMC's unit stiffness, and GMC's unit length. We excluded the magnetization profile of the GMC from the list of parameters, despite its consideration in other studies, because changing the magnetization profile of the magnetic continuum produces similar bending results to altering the external magnetic field.
% Considering changing the magnetization profile of the magnetic continuum will have a similar bending result as changing the external magnetic field, we remove the magnetization profile of GMC from the parameters list, even though it was studied in other studies. 
In general, increasing GMC's cross-section side length leads to an increase in the bending radius. Although the total mass of neodymium magnetic nanoparticles (NdFeB) also increases, the effect of the cross-section area remains dominant. Decreasing the stiffness of each section results in a smaller bending radius under the same magnetic field. Additionally, increasing the unit length of the rigid/soft sections generates a larger/smaller bending radius (Fig.~\ref{figure1}.C).

\subsection{Fabrication of GMC}

The thermal and physical properties of MIMMs \cite{c11} make it feasible to apply multiple fabrication methods to millimeter and sub-millimeter scale GMCs, including molding and 3D printing. In this study, we used the molding method to rapidly prototype GMCs with a graded stiffness structure and a simple shape design (Fig.~\ref{figure1}.D).

The preparation step starts with mixing two different stiffness PCL particles (150 $\mu m$, Ruixiang Plastics Inc.) and neodymium magnetic nanoparticles (5 $\mu m$, Magnequench Co., Ltd.). The ratios between two PCL particles are fixed in all bending experiments. We used a weight ratio of 2:1 (soft PCL/rigid PCL) for the bottom section (fixed end), a weight ratio of 5:1 for the middle section (transition), and a weight ratio of 6:1 for the top section (tip). These materials are selectively weighted and mixed in a glass petri dish. After mixing, the glass petri dish is placed on a heating platform (Yarun Inc.), and the mixed particles are heated and stirred by a glass rod for 5-10 minutes. The time required for this step varies with different weight ratios of mixed nanoparticles.

We used Ecoflex 00-30 (Smooth-On, USA) to create molds for continuums with different outer diameters (0.75 mm, 1 mm, and 1.25 mm). The silicone rubber mold is filled with the heated liquid mixture mentioned in the previous step, and a heated scraper is slowly moved from one side to the other to create a flat surface. The softened MC is quickly cooled to ambient temperature by placing the mold with the continuum in cold water for 1 minute. After cooling down, the MC can be easily demolded from the silicone rubber mold. The uniform MC is then divided into 10 mm sections for the final connecting process. Three sections of MCs with graded-changing stiffness are placed in the mold and reheated at intersections for merging. The same cooling and demolding processes are conducted to obtain a GMC.

% \clearpage

\section{Experiments and Simulations}

Bending is a fundamental performance of GMC, enabling the continuum robots to navigate tortuous pathways and execute specific tasks. In this study, we conducted bending experiments of GMCs with different design parameters. For real experiments, we varied the external magnetic field strength, GMC's cross-section, each section's unit length, and the weight percentage of NdFeB nanoparticles. When the weight percentage of NdFeB nanoparticles is altered, the Young's modulus of each continuum section changes, accompanied by a variation in magnetization strength. Consequently, this parameter has a more complex influence on GMC's bending behavior compared to changing unit stiffness.

To validate our numerical model in COMSOL, we repeated the previously mentioned bending tests in the numerical model, using the corresponding magnetic field strength, unit cross-section, and unit length. The simulation results were compared with the real bending test results for validation purposes. Additionally, we introduced one more independent factor to the numerical model: unit stiffness (Young's modulus), which is crucial to the final bending shape. This factor replaced the NdFeB weight ratio in bending experiments, as they have the same effect on stiffness, while the magnetization profile was input separately in the numerical model.
% We did not include this factor in the physical bending tests, as it is difficult to modify the stiffness during the molding fabrication process, whereas it was much easier in the COMSOL simulation. 
The results of the numerical model are plotted in the same figure as real bending tests, demonstrating the reliability of our proposed model.

\subsection{Bending Experiments}

\begin{table}[b]
\caption{GMC design parameters}
\label{table1}
\begin{center}
\resizebox{\linewidth}{!}{ % 调整表格宽度以适应文本宽度，高度按比例缩放
\begin{tabular}{c|c|c|c|c}
    \toprule
    % \multirow{2}{*}{No.} & \multirow{2}{*}{Cross-section} & \multirow{2}{*}{Unit length} & \multirow{2}{*}{NdFeB} & \multirow{2}{*}{Unit Stiffness}\\
    %  & side length & & weight ratio & (COMSOL)\\
    \multirow{1}{*}{No.}& \parbox{1.6cm}{\centering Cross-section \\Unit Length}& \parbox{1.6cm}{\centering Unit Length} & \parbox{1.6cm}{\centering NdFeB\\Weight Ratio} &\parbox{1.8cm}{\centering Unit Stiffness\\(COMSOL)}\\
    \midrule
    1 & 1.25 mm & 10-10-10 mm & 200\% & 20-15-10 MPa\\
    % \midrule
    2 & 1.0 mm & 10-10-10 mm & 200\% & 20-15-10 MPa\\
    % \midrule
    3 & 0.75 mm & 10-10-10 mm & 200\% & 20-15-10 MPa\\
    % \midrule
    4 & 1.0 mm & 10-10-10 mm & 150\% & 16-12-8 MPa\\
    % \midrule
    5 & 1.0 mm & 10-10-10 mm & 100\% & 14-10-7.5 MPa\\
    % \midrule
    6 & 1.0 mm & 20-5-5 mm & 200\% & 20-15-10 MPa\\
    % \midrule
    7 & 1.0 mm & 5-5-20 mm & 200\% & 20-15-10 MPa\\
    \bottomrule
    \end{tabular}
}
\end{center}
\end{table}

   \begin{figure}[h]
      \centering
      \includegraphics[width=\linewidth]{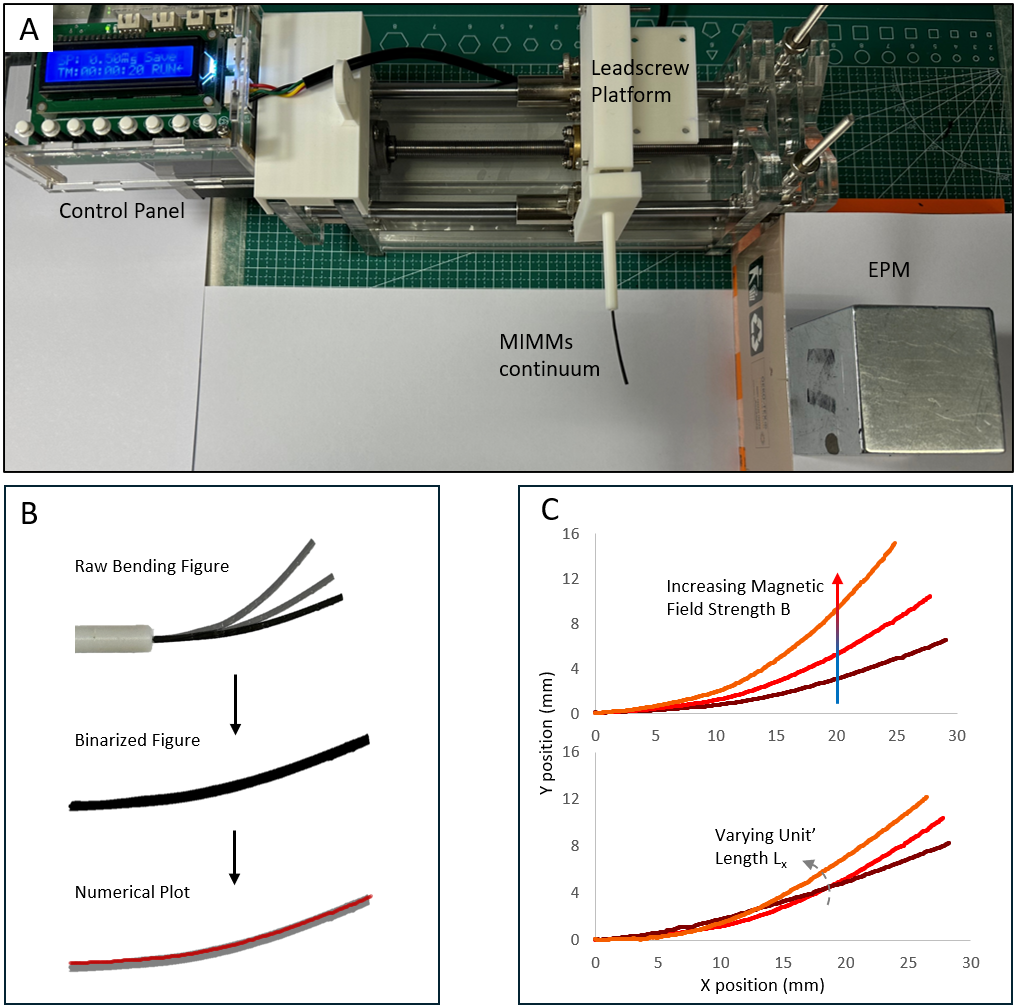}
      \caption{\textbf{Bending Tests.} (A) Experimental setup for GMC bending tests. (B) Processing steps from the raw image to data points. (C) Bending performances of a GMC, when changing the external magnetic field strength/unit length of each section.}
      \label{figure2}
   \end{figure}

   % [data plot is unclear and without xylabel]

In the bending experiments, we measured and recorded the shapes of GMCs under an EPM's magnetic field. The cubic EPM has a side length of 55 mm. The experimental setup included a motor-driven leadscrew, a control panel, an EPM, and a GMC (Fig~\ref{figure2}.A). A replaceable GMC was fixed to a 3D printed platform on the leadscrew, with its tip aligned with the N pole center of the EPM. Upon receiving a command from the control panel, the platform approached the EPM in 10 mm steps. The bending shapes of the GMC after each step were recorded by a camera from the top. 

To avoid sharp bending at the GMC's fixed end, we applied a graded-stiffness design strategy. Commercially available PCL powders with the same material properties but different initial stiffness were used. We prepared two PCL powders with different stiffness and mixed them in different weight ratios as the base material. As mentioned in the fabrication section, the GMC was molded and behaved in a circular curvature under an EPM's magnetic field. Specifically, we used 7 different GMCs with varying parameters for the bending experiments (Table 1). The cross-section side lengths were 1.25 mm, 1.0 mm, and 0.75 mm. The unit lengths of the three sections (bottom, middle, and top sections) were 10-10-10 mm, 20-5-5 mm, and 5-5-20 mm. The weight ratios of NdFeB were 200\%, 150\%, and 100\%. Each GMC was placed at three adjacent positions (with a 10 mm step) and formed bending shapes under three different EPM's magnetic field strengths. These GMCs' bending performances provided ten sets of comparison results for specific parameters.

We used Wolfram Mathematica (Wolfram Research, Inc.) to process the images captured by the camera (Fig~\ref{figure2}.B). A white paper was placed beneath the curving GMC to generate a sharp contrast between a black GMC and the white background. The raw figure was binarized in the following step, and the position information of each point on the continuum's side edge was recorded. This position data was used to present the final bending shape.
% The final bending shape was presented using this position information. 
The influence of changing the external magnetic field or unit length on the bending shape is evident. As EPM's magnetic field strength or GMC top section's unit length increases, the bending radius of the continuum decreases significantly (Fig.~\ref{figure2}.C). 

% \begin{itemize}

% \item Use either SI (MKS) or CGS as primary units. (SI units are encouraged.) English units may be used as secondary units (in parentheses). An exception would be the use of English units as identifiers in trade, such as Ò3.5-inch disk driveÓ.
% \item Avoid combining SI and CGS units, such as current in amperes and magnetic field in oersteds. This often leads to confusion because equations do not balance dimensionally. If you must use mixed units, clearly state the units for each quantity that you use in an equation.
% \item Do not mix complete spellings and abbreviations of units: ÒWb/m2Ó or Òwebers per square meterÓ, not Òwebers/m2Ó.  Spell out units when they appear in text: Ò. . . a few henriesÓ, not Ò. . . a few HÓ.
% \item Use a zero before decimal points: Ò0.25Ó, not Ò.25Ó. Use Òcm3Ó, not ÒccÓ. (bullet list)

% \end{itemize}

\subsection{COMSOL Simulations}

\begin{figure}[t]
      \centering
      \includegraphics[width=\linewidth]{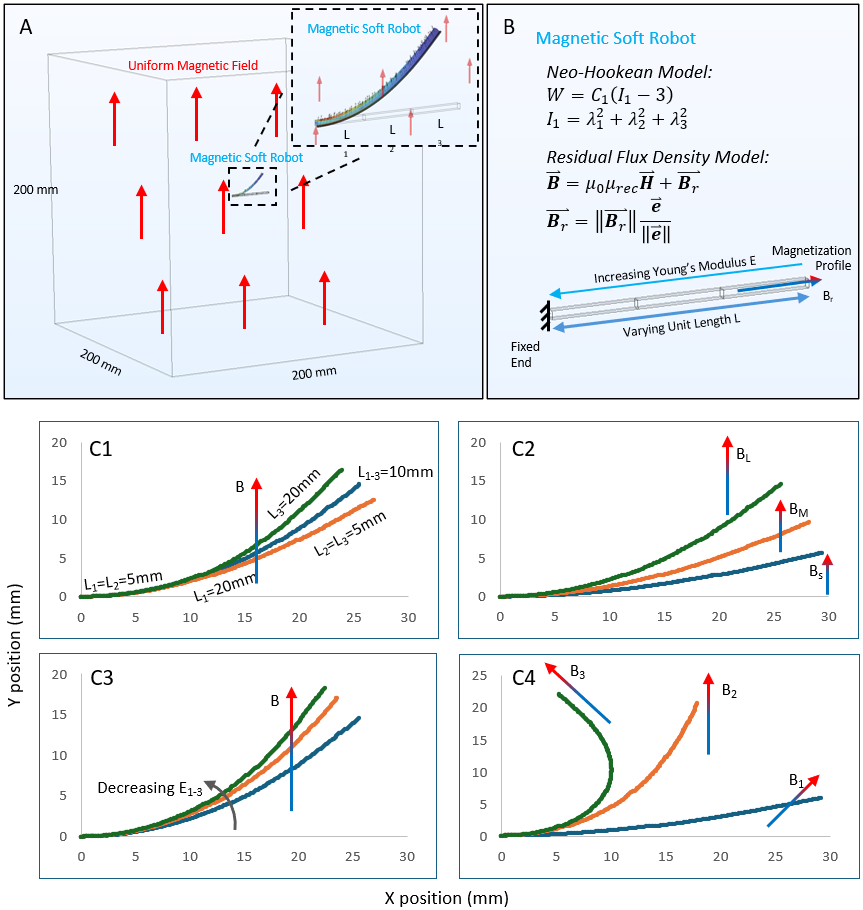}
      \caption{\textbf{COMSOL Simulations of the GMC.} (A) Environmental setup for a uniform magnetic field. (B) Magnetic Soft Robot model for the simulation of the GMC. (C1) Numerical results with varying unit lengths. (C2) Numerical results with increasing magnetic field strengths. (C3) Numerical results with decreasing unit stiffness. (C4) Numerical results with different magnetic field angles.}
      \label{figure3}
   \end{figure}

We constructed a numerical model using COMSOL Multiphysics (COMSOL, Inc.) to validate the GMC's bending behaviors. To simplify the hyperelastomer-magnetic coupling model for the GMC and reduce simulation time, a uniform magnetic field is applied. The study on the EPM magnetic field \cite{c12} demonstrated with Equation 1 that the magnetic field strength decreases significantly as the distance from the EPM's N pole center increases. This phenomenon is consistent with our GMC bending experiments, and the GMC's harder bottom section further reduces the influence of the magnetic field. As a result, it is reasonable to apply this hypothesis to our numerical model using a uniform magnetic field (Fig.~\ref{figure3}.A). The uniform magnetic field strength was set to the same value as we measured in the bending experiments. We analyzed the values of magnetic field strength using the following equations.

% \begin{flalign}
% &H_Z=\frac{B_r}{4\pi\mu_0}& 
% \nonumber
% \end{flalign}

% \begin{flalign}
% &\times\int_{X_0+(X_m/2)}^{X_0-(X_m/2)}\int_{Y_0+(Y_m/2)}^{Y_0-(Y_m/2)}\left(\frac{Z_0 - \frac{Z_m}{2}}{r_N^{'3}}-\frac{Z_0+\frac{Z_m}{2}}{r_S^{'3}}\right)dy'dx', &
% \end{flalign}

\begin{equation}
    H_Z=\frac{B_r}{4\pi\mu_0}\int_{X_0+(X_m/2)}^{X_0-(X_m/2)}\int_{Y_0+(Y_m/2)}^{Y_0-(Y_m/2)}\Delta r dy'dx',
\end{equation}
\begin{equation}
    \Delta r = \frac{Z_0 - \frac{Z_m}{2}}{r_N^{'3}}-\frac{Z_0+\frac{Z_m}{2}}{r_S^{'3}},
\end{equation}

where $B_r$ is the flux density, $X_0, Y_0, Z_0$ are the coordinates of the GMC tip in our study case, $X_m = Y_m = Z_m = 55 mm$ for our EPM, and ${r_N^{'3}},{r_S^{'3}}$ represent the distances from the continuum tip to any small elements on the surface of N/S poles. ${r_N^{'3}},{r_S^{'3}}$ can be calculated as shown below.

\begin{equation}
r'_N = \sqrt{(X_0 - x)^2+(Y_0 + y)^2+\left(Z_0-\frac{Z_m}{2}\right)^2}
\end{equation}
\begin{equation}
r'_S = \sqrt{(X_0 - x)^2+(Y_0 - y)^2+\left(Z_0+\frac{Z_m}{2}\right)^2}
\end{equation}

\begin{equation}
\vec{B} = \mu_0\vec{H}
\end{equation}

We detected a magnetic field strength of 38 mT, when the probe of a Tesla meter is 70 mm away from N pole center; a magnetic field strength of 50 mT, when the probe is 60 mm away from N pole center; a magnetic field strength of 66 mT, when the probe is 50 mm away from N pole center. Based on Eqs. 1-4, the flux densities $B_r$ are equal to 1.3578 T, 1.3050 T, 1.2938 T, respectively. These values are acceptable, as the typical $B_r$ for an NdFeB EPM ranges from 1.2 T to 1.4 T. In the numerical model, the magnetic field strength was defined by specifying a magnetic flux density, using the previously detected values.

   % [data plot need xy label]

For the GMC (magnetic soft robot in the numerical model), we assigned the Neo-Hooken Model (Fig.~\ref{figure3}.B), which provides accurate hyperelastic descriptions when the strain is lower than 100\%. A fixed support was applied to the bottom section end as a boundary condition. We also restricted the movements of the GMC in the X and Y directions. The dynamic mesh technique was applied to accommodate mesh deformations during the bending performances. The residual flux density was applied to describe the magnetization profile of the GMC (Fig.~\ref{figure3}.B). We measured the magnetization profile of continuum No. 1-7 in the following order: 25.07 mT, 20.07 mT. 14.59 mT, 19.71 mT, 16.18 mT, 19.19 mT, 21.82 mT. These parameters were input into the magnet soft robot model to characterize GMC's properties. In addition, we applied the Maxwell tensor at the top section end of the GMC for interaction with the magnetic field.
   \begin{figure*}[h!]
      \centering
      \includegraphics[width=0.75\textwidth]{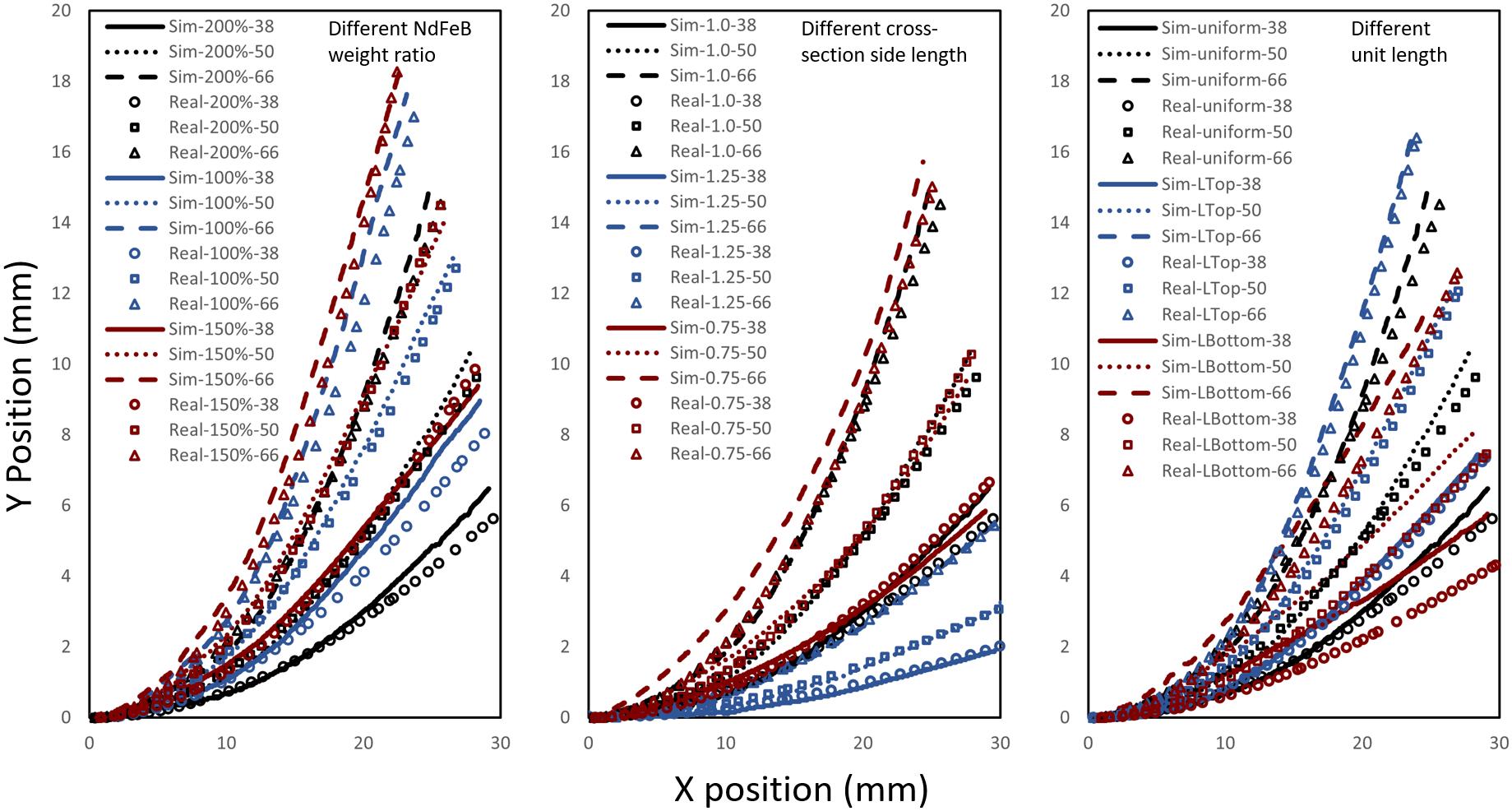}
      \caption{\textbf{Comparison Results between Real Bending Experiments and COMSOL Simulations.}}
      \label{figure4}
   \end{figure*}
   
To apply the graded-stiffness design strategy to the GMC model, we divided the magnet soft robot into three sections. The unit length of each section was assigned based on continuum parameters from Table 1. The only unknown parameter remaining at this stage was the stiffness of each section. According to our previous research on MIMMs \cite{c11}, the Young's modulus range of MIMMs is approximately 8-50 MPa. While keeping all the other parameters fixed and decreasing stiffness from the bottom to the top section, we adjusted each section's Young's modulus to fit the simulation outputs with the real bending performances. As a result, we applied 20 MPa, 15 MPa, 10 MPa to the bottom, middle, top section of 200\% NdFeB weight ratio GMCs; 16 MPa, 12 MPa, 8 MPa to the bottom, middle, top section of 150\% NdFeB weight ratio GMCs; 14 MPa, 10 MPa, 7.5 MPa to the bottom, middle, top section of 100\% NdFeB weight ratio GMCs. The bending performance differences were visualized through simulation by modifying each parameter (Fig.~\ref{figure3}.C).

\subsection{Results Validation}
   % \begin{figure}[b]
   %    \centering
   %    \includegraphics[width=\linewidth]{Figures/Figure4_v3.png}
   %    \caption{Comparison Results between Real Bending Experiments and COMSOL Simulation}
   %    \label{figure4}
   % \end{figure}
   
The bending shape of the GMC is principally determined by one external factor: magnetic field strength, and three internal factors: cross-section side length, unit stiffness (weight-ratio of NdFeB in bending experiments), and unit length. We categorized our study cases based on these three internal factors and plotted both real and numerical results in the same figure (Fig.~\ref{figure4}). 

In the left plot (Fig.~\ref{figure4}), only the NdFeB weight ratio and the external magnetic field are varied. The influence of the NdFeB weight-ratio is complex since it simultaneously changes unit stiffness and continuum magnetization strength. When the ratio increases from 100\% to 150\%, the magnetization increase is dominant, whereas from 150\% to 200\%, the stiffness hardening becomes more pronounced. In the middle plot (Fig.~\ref{figure4}), the GMC's cross-section side length and the external magnetic field strength are varied. We found that decreasing the cross-section side length helps the GMC to curve with a smaller bending radius, even though it reduces magnetization strength at the same time. In the right plot (Fig.~\ref{figure4}), the unit length of each section is altered (while keeping the total length constant) and the external magnetic field strength is varied. A longer bottom section (harder part) results in a larger bending radius, whereas a longer top section has the opposite effect.

In all 27 sets of comparisons between real bending experiment results and simulations, the bending curves are similar and exhibit the same trend when a single parameter is modified. We reasonably predict that, with any valid parameter inputs (cross-section side length, unit stiffness, unit length, and external magnetic field strength), our numerical COMSOL model can precisely predict the bending behavior of GMCs.

\section{Expansion Model Training}

\begin{figure}[t]
      \centering
      \includegraphics[width=0.9\linewidth]{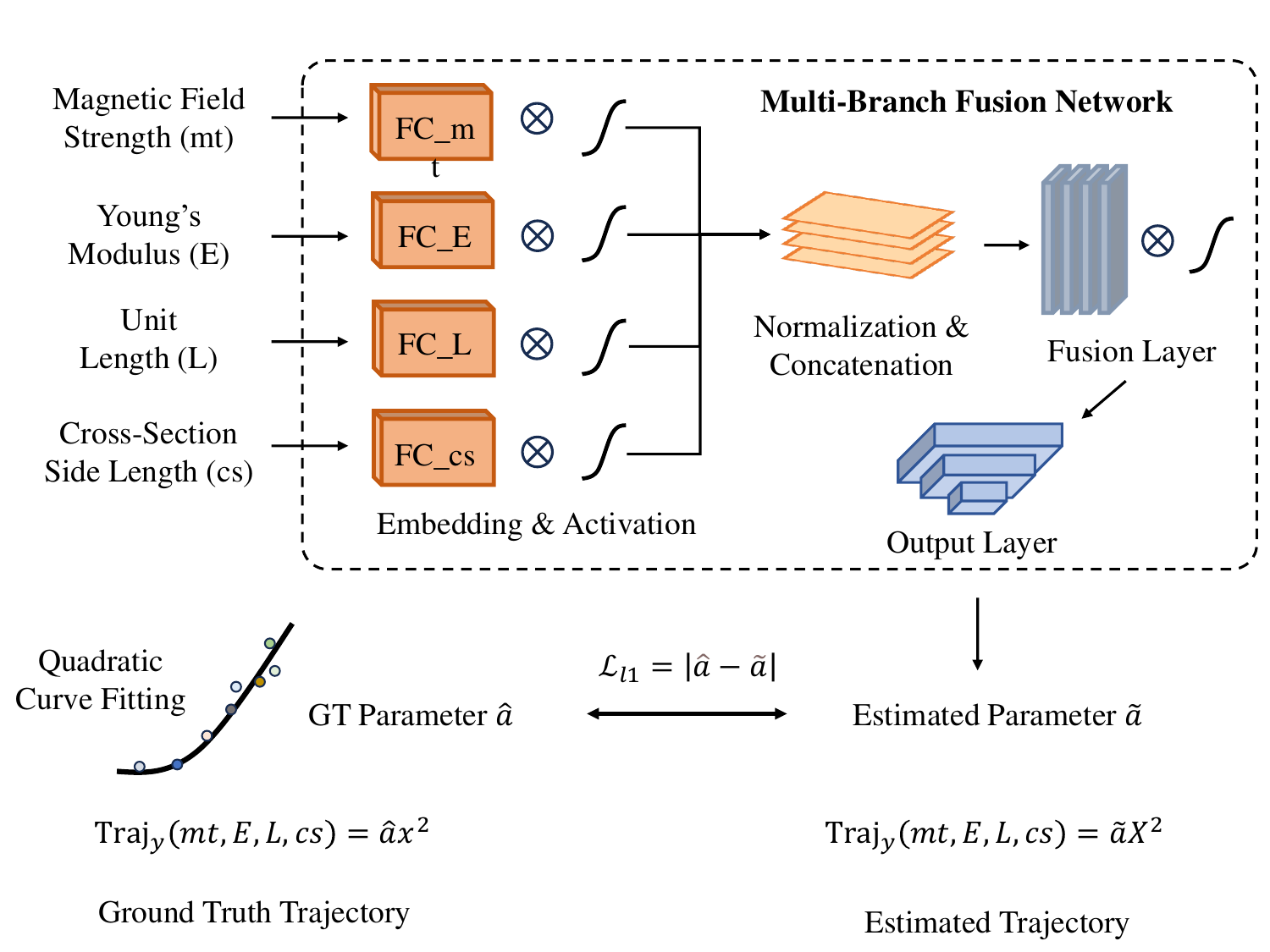}
      \caption{\textbf{Overview of the Expansion Estimation Network.} We apply a Multi-Branch for Expansion Trajectory Estimation.}
      \label{figure5}
   \end{figure}

The simulation process was generally slow with multiple input parameters in the case of GMCs (10-20 minutes per case). To generate precise predictions, we had to use parameterized analysis with a gradual increase in magnetic field strength, which made the entire process more time-consuming. As a result, we trained a model to efficiently expand our GMCs bending prediction library (Fig.~\ref{figure5}).

% We took the COMSOL simulation result as the ground truth. Since the bending performance is the result of magnetic force in one direction, we could fit the bending shape with the quadratic curve and simplify the curve as one ground truth parameter $\widehat{\alpha}$. We had 4 inputs to the model, including external magnetic field strength (mt), unit Young's modulus (E), unit length (L), and cross-section side length (cs). A multi-branch fusion network was applied to generate estimated parameter $\widetilde{\alpha}$. In detail, four input parameters would be embedded and activated separately, and then they would be normalized and concatenated. After going through a fusion layer and an output layer, the estimated parameter $\widetilde{\alpha}$ would be obtained. $L_1$ loss was applied to get the error between ground truth and estimation.
% 一鸣，这一段这个fusion网络怎么描述你帮我改改？并不是太懂%\

The proposed Multi-branch fusion network $\mathcal{F}$ first includes a four-branch embedding module with four fully connected layers: FC\_mt, FC\_E, FC\_L, and FC\_cs, corresponding to the magnetic field, Young's modules, unit length, and cross-section side length, respectively. The input embeddings are then normalized and concatenated for feature fusion. Finally, we generate the output estimation $\mathcal{F}(mt, E, L, cs) = \tilde{a}$ by the output layer. We assume the expansion of the GMC follows the morphology of the quadratic curve and propose to model the expansion with the following equation:
\begin{equation}
    \text{Traj}_y (mt, E, L, cs) = \tilde{a}x^2,
\end{equation}
where $(x, y)$ are the position of the 2D coordinate. $mt, E, L, cs$ represent the magnetic field strength, Young's modulus, unit length, and cross-section side length. We simplified the second and third elements of the quadratic function as 0 by assuming the starting point is flat and at the origin point $(x,y)=(0, 0)$. The model was optimized with L1 loss: $\mathcal{L} = \vert \hat{a}-\tilde{a}\vert$, where $\hat{a}$ is the ground truth parameter obtained from curve fitting using the ground truth points. We adopted Adam optimizer with a learning rate of $10^{-3}$.

\begin{figure}[t]
      \centering
      \includegraphics[width=\linewidth]{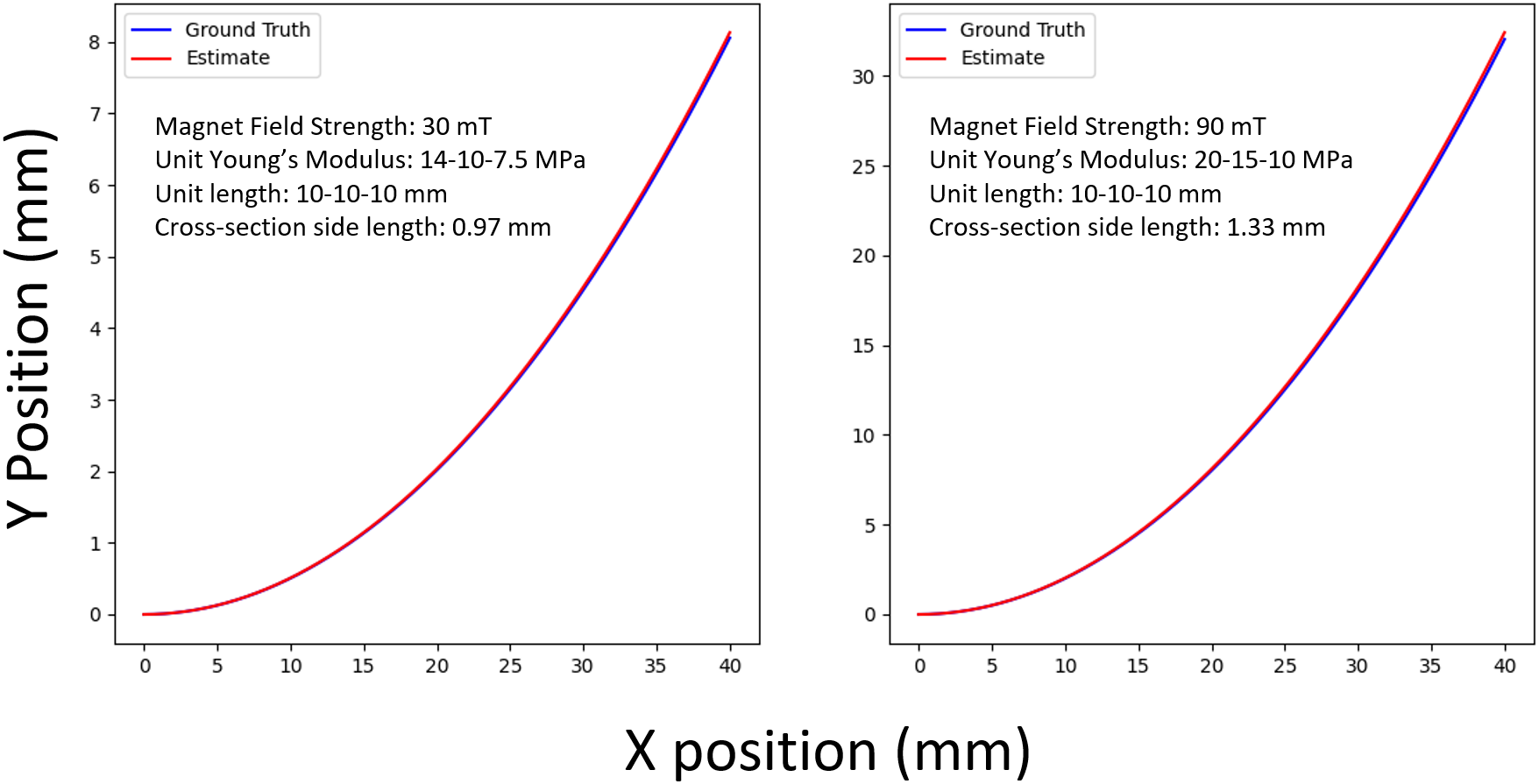}
      \caption{\textbf{Expansion Estimation Results.} Our model estimates a high-accuracy expansion trajectory, with a MSE of $2.089\times10^{-7}$.}
      \label{network_estimation}
   \end{figure}

We trained the expansion model using 7 categories of GMC parameters (Table 1). Each category covered an external magnetic field strength ranging from 10 mT to 120 mT, with increments of 10 mT. To achieve a more precise model, we remeasured the cross-section of the GMC. Three cross-section side lengths were measured at three points: the two ends and the middle of the continuum. We used the average of these three measurements as the new cross-section side length: 0.8 mm, 0.97 mm, and 1.30 mm (for both COMSOL simulation and expansion model training). A total of $7\times 11$ sets of point sets are used for training. The remaining $7\times 1$ sets of points are split from the dataset for testing. Our model achieves a result of MSE $2.089\times10^{-7}$ for the unseen testing cases. Qualitative results are shown in Fig.~\ref{network_estimation}, where the estimated expansion trajectory aligns well with the testing ground truth. These results indicate that the proposed model is highly efficient and accurate in predicting the GMC's bending behaviors. Based on the model, we are able to create an extensive library of GMCs bending behaviors with reasonable design parameters, enabling better decisions for specific bending requirements.

\section{Conclusions and Future Work}

In this work, we conducted bending tests of the GMC under an EPM magnetic field. The graded-stiffness design strategy avoids sharp bending at the fixed end of the GMC and produces a more circular bending curve, which is preferred in navigation tasks. We also constructed a numerical model for GMCs' bending performance and validated the numerical results with real bending experiments. Furthermore, we trained an expansion model based on numerical results to accurately and efficiently predict the bending behaviors of GMCs. The well-trained expansion model allows us to generate a large library of GMCs' bending information with four different input parameters.

In COMSOL Multiphysics, we used a uniform magnetic field to simulate the real environment. Although the bending simulation outputs of GMCs are close to the real experiments, there is a gap between the simulated uniform field and the EPM magnetic field. In the future, we can work on simulating the EPM environment to create a more accurate model. Last but not least, the expansion model could be further developed into a backward conjecture model. This means that given a desired bending performance or bending radius, the model will provide suggestions for feasible design parameters.

\addtolength{\textheight}{-12cm}   % This command serves to balance the column lengths
                                  % on the last page of the document manually. It shortens
                                  % the textheight of the last page by a suitable amount.
                                  % This command does not take effect until the next page
                                  % so it should come on the page before the last. Make
                                  % sure that you do not shorten the textheight too much.

%%%%%%%%%%%%%%%%%%%%%%%%%%%%%%%%%%%%%%%%%%%%%%%%%%%%%%%%%%%%%%%%%%%%%%%%%%%%%%%%

%%%%%%%%%%%%%%%%%%%%%%%%%%%%%%%%%%%%%%%%%%%%%%%%%%%%%%%%%%%%%%%%%%%%%%%%%%%%%%%%

%%%%%%%%%%%%%%%%%%%%%%%%%%%%%%%%%%%%%%%%%%%%%%%%%%%%%%%%%%%%%%%%%%%%%%%%%%%%%%%%
% \section*{APPENDIX}

% Appendixes should appear before the acknowledgment.

% \section*{ACKNOWLEDGMENT}

% The preferred spelling of the word ÒacknowledgmentÓ in America is without an ÒeÓ after the ÒgÓ. Avoid the stilted expression, ÒOne of us (R. B. G.) thanks . . .Ó  Instead, try ÒR. B. G. thanksÓ. Put sponsor acknowledgments in the unnumbered footnote on the first page.

%%%%%%%%%%%%%%%%%%%%%%%%%%%%%%%%%%%%%%%%%%%%%%%%%%%%%%%%%%%%%%%%%%%%%%%%%%%%%%%%

% References are important to the reader; therefore, each citation must be complete and correct. If at all possible, references should be commonly available publications.


\begin{thebibliography}{99}

\bibitem{c1} Kim, Y., Parada, G. A., Liu, S., \& Zhao, X. (2019). Ferromagnetic soft continuum robots. Science robotics, 4(33), eaax7329.
\bibitem{c2} Liu, X., Wang, L., Xiang, Y., Liao, F., Li, N., Li, J., ... \& Zang, J. (2024). Magnetic soft microfiberbots for robotic embolization. Science Robotics, 9(87), eadh2479.
\bibitem{c3} Lussi, J., Mattmann, M., Sevim, S., Grigis, F., De Marco, C., Chautems, C., ... \& Nelson, B. J. (2021). A submillimeter continuous variable stiffness catheter for compliance control. Advanced Science, 8(18), 2101290.
\bibitem{c4} Su, S., \& Ren, H. (2025). Robust magnetic tracking and navigation in robotic surgery. \textit{Handbook of Robotic Surgery}, 69-80. 
\bibitem{c5} Limpabandhu, C., Hu, Y., Ren, H., Song, W., \& Tse, Z. T. H. (2023). Actuation technologies for magnetically guided catheters. Minimally Invasive Therapy \& Allied Technologies, 32(4), 137-152. 
\bibitem{c6} Cao, Y., Yang, Z., Hao, B., Wang, X., Cai, M., Qi, Z., ... \& Zhang, L. (2023). Magnetic continuum robot with intraoperative magnetic moment programming. Soft Robotics, 10(6), 1209-1223.
\bibitem{c7} Park, J., Kee, H., \& Park, S. (2024). Workspace Expansion of Magnetic Soft Continuum Robot using Movable Opposite Magnet. IEEE Robotics and Automation Letters.
\bibitem{c8} Chautems, C., Tonazzini, A., Boehler, Q., Jeong, S. H., Floreano, D., \& Nelson, B. J. (2020). Magnetic continuum device with variable stiffness for minimally invasive surgery. Advanced Intelligent Systems, 2(6), 1900086.
\bibitem{c9} Wang, L., Zheng, D., Harker, P., Patel, A. B., Guo, C. F., \& Zhao, X. (2021). Evolutionary design of magnetic soft continuum robots. Proceedings of the National Academy of Sciences, 118(21), e2021922118.
\bibitem{c10} Lloyd, P., Thomas, T. L., Venkiteswaran, V. K., Pittiglio, G., Chandler, J. H., Valdastri, P., \& Misra, S. (2023). A magnetically-actuated coiling soft robot with variable stiffness. IEEE Robotics and Automation Letters, 8(6), 3262-3269.
\bibitem{c11} Yang, Y., Yuan, S., \& Ren, H. (2024). Reversible elastomer–fluid transitions for metamorphosic robots. Advanced Functional Materials, 34(18), 2311981.
\bibitem{c12} Nishimura, K. (2021). Three‐dimensional array of strong magnetic field by using cubic permanent magnets. Electrical Engineering in Japan, 214(1), 18-25.
% \bibitem{c12} R. W. Lucky, ÒAutomatic equalization for digital communication,Ó Bell Syst. Tech. J., vol. 44, no. 4, pp. 547Ð588, Apr. 1965.
% \bibitem{c13} S. P. Bingulac, ÒOn the compatibility of adaptive controllers (Published Conference Proceedings style),Ó in Proc. 4th Annu. Allerton Conf. Circuits and Systems Theory, New York, 1994, pp. 8Ð16.
% \bibitem{c14} G. R. Faulhaber, ÒDesign of service systems with priority reservation,Ó in Conf. Rec. 1995 IEEE Int. Conf. Communications, pp. 3Ð8.
% \bibitem{c15} W. D. Doyle, ÒMagnetization reversal in films with biaxial anisotropy,Ó in 1987 Proc. INTERMAG Conf., pp. 2.2-1Ð2.2-6.
% \bibitem{c16} G. W. Juette and L. E. Zeffanella, ÒRadio noise currents n short sections on bundle conductors (Presented Conference Paper style),Ó presented at the IEEE Summer power Meeting, Dallas, TX, June 22Ð27, 1990, Paper 90 SM 690-0 PWRS.
% \bibitem{c17} J. G. Kreifeldt, ÒAn analysis of surface-detected EMG as an amplitude-modulated noise,Ó presented at the 1989 Int. Conf. Medicine and Biological Engineering, Chicago, IL.
% \bibitem{c18} J. Williams, ÒNarrow-band analyzer (Thesis or Dissertation style),Ó Ph.D. dissertation, Dept. Elect. Eng., Harvard Univ., Cambridge, MA, 1993. 
% \bibitem{c19} N. Kawasaki, ÒParametric study of thermal and chemical nonequilibrium nozzle flow,Ó M.S. thesis, Dept. Electron. Eng., Osaka Univ., Osaka, Japan, 1993.
% \bibitem{c20} J. P. Wilkinson, ÒNonlinear resonant circuit devices (Patent style),Ó U.S. Patent 3 624 12, July 16, 1990. 






\end{thebibliography}
\end{document}